\documentclass{article}

\usepackage{arxiv}          

\usepackage[square,numbers,comma]{natbib} 
\usepackage[utf8]{inputenc} 
\usepackage[T1]{fontenc}    
\usepackage{hyperref}       
\usepackage{url}            
\usepackage{booktabs}       
\usepackage{multirow}       
\usepackage{amsfonts}       
\usepackage{nicefrac}       
\usepackage{microtype}      
\usepackage{xcolor}         
\usepackage{amsmath}
\usepackage{graphicx}
\usepackage{tikz}
\usetikzlibrary{arrows.meta, positioning}
\usepackage{float}

\title{When Does Domain Adaptation Help on Physical Vibration Sensors?
A Held-Out-Bearing Study of Neural-Operator and Convolutional Models}

\author{
  Kumbha Nagaswetha\\
  India\\
  \texttt{nagaswethak@.iisc.ac.in}\\
  \And
  Rabi Pathak\\
  India\\
  \texttt{rabipathak@iisc.ac.in}\\
}

\begin{document}

\maketitle

\begin{abstract}
Diagnosing rolling-element bearing faults from vibration is a canonical
physical-sensing task and a widely used benchmark for domain adaptation under
operating-condition shift. Accuracies above 99 percent are commonly reported,
but under evaluation splits that place the same physical bearing in both training
and test. We revisit the task under a held-out-bearing protocol, assigning every
bearing unit entirely to either the training or the test set, and find that
source-only transfer is far weaker than such numbers suggest: on a change of
shaft speed it reaches only $0.36$, against a target-supervised ceiling of
$0.97$. We then study what governs transfer. Treating computed order tracking, a
shaft-angle resampling that places fault frequencies at fixed shaft orders
independent of running speed, as a controlled change of representation, we find
that a Fourier Neural Operator raises source-only transfer from $0.36$ to $0.61$
on the speed shift, where the fault peaks move, while a convolutional network of
matched feature dimension stays near chance in both representations. The
representation also decides whether unsupervised alignment can work: with the
same normalized RBF-MMD loss and no target labels, the operator reaches $0.71$ in
the frequency domain but $0.95$ in the order domain, within $0.02$ of the
target-supervised ceiling and above $0.86$ on every held-out bearing fold. Once
the representation is right, a small label budget adds little. These results
indicate that, for this task, the input representation rather than the alignment
method decides whether adaptation helps. A second dataset, whose held-out units
are fault diameters rather than bearings, shows that the same protocol exposes
failures that even a target-supervised model cannot avoid.
\end{abstract}


\section{Introduction}
\label{sec:intro}

Rolling-element bearings are among the most common failure points in rotating
machinery, and monitoring their health from vibration signals is a standard task
in predictive maintenance~\citep{randall2011}. In practice, a diagnostic model is
trained on data collected under one operating condition, such as a particular
shaft speed or load, but must run under other conditions for which labeled data
are scarce or unavailable. This mismatch between training and deployment
conditions is a domain adaptation (DA) problem, and it has been studied
extensively on public bearing datasets~\citep{lessmeier2016, smith2015, zhao2020}.

Much of this literature reports very high accuracy, frequently above 99
percent~\citep{zhao2020}. These results, however, are usually obtained under
evaluation splits that partition individual vibration windows, so that windows
from the same physical bearing appear in both the training and the test set.
Because each bearing carries its own manufacturing and mounting signature, such
splits allow a model to recognize the specific unit rather than the fault type,
and the reported accuracy can reflect this recognition as much as genuine
transfer across conditions~\citep{hendriks2022}. From the published numbers
alone, it is therefore difficult to say how well these methods adapt when the
test bearing has never been seen.

We revisit the task under a held-out-bearing protocol, in which each physical
bearing unit is assigned in its entirety to either the training or the test set.
Under this protocol, source-only transfer is markedly weaker than the standard
splits suggest, which raises a more basic question than which alignment method
performs best: what determines whether adaptation succeeds at all? We approach
this question through the physics of the signal. A local defect produces impacts
at frequencies that are fixed multiples of the shaft rotation rate, so a change
in shaft speed moves these fault frequencies along the frequency axis. Computed
order tracking removes this movement by resampling the signal against shaft angle
rather than time, which places the fault frequencies at fixed positions
regardless of speed. We use order tracking as a controlled change of input
representation and compare two architectures of matched feature dimension: a
Fourier Neural Operator~\citep{li2021fno}, whose computation is carried out in the
frequency domain and is therefore sensitive to where spectral energy sits, and a
convolutional network, which responds to local patterns in time. More broadly,
the lesson is that encoding a known invariance of the sensor - here, the
periodicity of shaft rotation - into the coordinate system before learning can
matter more than the choice of adaptation objective, which is relevant to any
sensor mounted on a periodic mechanical system.

Our contributions are the following.
\begin{itemize}
\item We evaluate bearing fault domain adaptation under a held-out-bearing
  protocol, and show that source-only transfer is substantially lower than the
  values reported under window-level splits.
\item Using order tracking as a controlled intervention, we show that the neural
  operator improves its source-only transfer on the shift that moves the fault
  frequencies, and that the same representation is what allows unsupervised
  alignment to work at all, whereas a convolutional network of matched feature
  dimension is essentially unaffected. This isolates the input representation,
  rather than the alignment loss, as the factor that determines whether
  adaptation helps.
\item We provide a systematic comparison across two architectures at matched
  feature dimension, five alignment choices, and six target-label budgets, with
  paired statistics over nine held-out-bearing runs per configuration, and show
  that a small amount of labeled target data recovers most of the remaining gap.
  On a second dataset we show that the protocol also exposes unseen-fault-geometry
  failures that a target-supervised model shares.
\end{itemize}

\section{Related Work}
\label{sec:related}
\paragraph{Data-driven bearing fault diagnosis and domain adaptation:}
Deep networks, and one-dimensional convolutional networks in particular, are now
standard for classifying bearing faults directly from vibration signals, and a
large body of work adapts them across operating conditions when labeled target
data are limited~\citep{zhao2020}. Common alignment losses, including correlation
alignment~\citep{sun2016coral}, domain-adversarial training~\citep{ganin2016dann},
and maximum mean discrepancy~\citep{gretton2012mmd}, have all been applied in this
setting to reduce the distribution gap between source and target conditions, and
benchmark studies collect such methods on public datasets and report their
transfer accuracy~\citep{zhao2020}.

\paragraph{Order tracking and the order domain:}
Removing the effect of running speed by working in the order domain rather than
the frequency domain is well established in diagnostics; envelope analysis and
order tracking are standard tools for extracting speed-invariant fault
features~\citep{randall2011}. This idea has also been combined with deep learning:
order-tracked signals have been classified with one-dimensional convolutional
networks, with improved transfer across speeds~\citep{ji2021order}; closer to
our setting, an order-based representation has been paired with an explicit domain
adaptation network~\citep{xu2023tco}, and recent work combines order-frequency
pre-processing with source-free test-time adaptation~\citep{jeong2025sfda}. Our
order representation is therefore not
itself novel. We use it not as a proposed method but as a controlled intervention
that separates the effect of the input representation from the effect of the
alignment loss.

\paragraph{Neural operators and spectral models:}
The Fourier Neural Operator~\citep{li2021fno} carries out its main computation in
the frequency domain, retaining a limited set of Fourier modes, which gives it an
inductive bias toward global spectral structure. We use this bias directly:
because bearing fault information appears as peaks at particular frequencies, a
model that operates in the frequency domain should be sensitive to where those
peaks sit, and hence to whether a shift moves them. We contrast it with a
convolutional network, whose bias is toward local patterns in time.

\paragraph{Evaluation practice:}
Reported transfer accuracy is sensitive to how the data are split. When
individual vibration windows are partitioned without keeping each physical
bearing entirely within one split, a model can learn to recognize the specific
unit rather than the fault type~\citep{hendriks2022}, and a recent study across
three public datasets reaches the same conclusion, arguing for bearing-wise
partitioning as the default and noting the residual leakage of splitting a
single healthy recording~\citep{vieira2025realistic}. We adopt a held-out-bearing
protocol that avoids this, and examine how the conclusions change.

\begin{figure}[t]
\centering
\resizebox{\linewidth}{!}{%
\begin{tikzpicture}[
  >=Stealth, node distance=7mm, font=\small,
  box/.style={draw, rounded corners, align=center, text width=20mm, minimum height=13mm, inner sep=3pt},
  a/.style={fill=blue!5}, b/.style={fill=orange!12}, c/.style={fill=green!8}, d/.style={fill=gray!8}
]
\node[box,a] (raw) {Raw\\vibration\\[2pt]\scriptsize 64 / 12\,kHz};
\node[box,a, right=of raw] (prep) {Envelope\\preprocessing\\[2pt]\scriptsize band-pass, Hilbert, decimate};
\node[box,b, right=of prep] (rep) {Representation\\[2pt]\scriptsize Hz ($L{=}2000$)\\ or order ($L{=}1920$)};
\node[box,c, right=of rep] (model) {FNO \, / \, CNN\\[2pt]\scriptsize matched feature dim.};
\node[box,d, right=of model] (da) {DA training\\[2pt]\scriptsize source $+\,f$ target labels, alignment};
\node[box,d, right=of da] (eval) {Held-out-bearing\\evaluation\\[2pt]\scriptsize 3 classes};
\foreach \x/\y in {raw/prep, prep/rep, rep/model, model/da, da/eval} \draw[->] (\x) -- (\y);
\end{tikzpicture}}
\caption{Overview. Raw vibration is reduced to a common envelope, expressed in
either the frequency (Hz) or the order domain, classified by a Fourier Neural
Operator or a convolutional network of the same feature size, and adapted across
operating conditions under a held-out-bearing protocol.}
\label{fig:pipeline}
\end{figure}
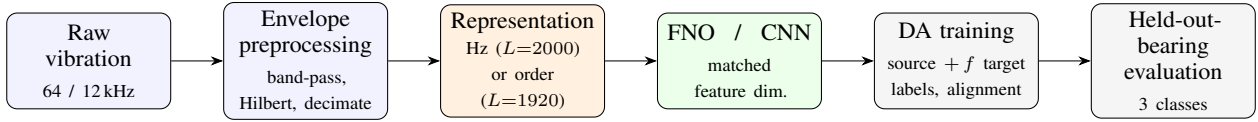

\section{Setup and Evaluation Protocol}
\label{sec:setup}

\paragraph{Datasets:}
Our main study uses the Paderborn University (PU) dataset~\citep{lessmeier2016},
which records a type-6203 bearing at 64~kHz under four operating conditions that
differ in shaft speed, load torque, and radial force; we take the 1500~rpm
($25$~Hz shaft rate) condition as the source and the 900~rpm ($15$~Hz, speed),
reduced-torque, and reduced-force conditions as targets. Each class is
represented by three physical bearings: healthy K001--K003, outer-race
KA01/KA03/KA04, and inner-race KI01/KI03/KI04. Of these, KA01, KA03, KI01 and
KI03 carry artificial damage, whereas KA04 and KI04 carry real damage from
accelerated-lifetime tests~\citep{lessmeier2016}; the third fold therefore trains
on artificial damage and tests on real damage (Section~\ref{sec:discussion}). As a
secondary dataset we use the
Case Western Reserve University (CWRU) data~\citep{smith2015}, which records a
type-6205 drive-end bearing at 12~kHz under four motor loads across which the
shaft rate changes by only $3.7$ percent ($29.95$ to $28.83$~Hz); we take the
lightest load as the source and the other three as targets. CWRU has a single
recording per fault type and diameter and a single healthy recording per load,
which constrains the protocol below. In both datasets we classify
three states: healthy, outer-race fault, and inner-race fault. For CWRU we keep
the outer-race faults at the six-o'clock position and exclude ball faults, so
that the label set matches PU.

\paragraph{Envelope representation:}
Both datasets are reduced to a common one-second envelope signal at 2~kHz. A
local defect excites a high-frequency structural resonance once per impact, so
the fault information is carried by the amplitude modulation of that resonance
rather than by a single spectral line. We therefore band-pass each window around
the resonance (2 to 6~kHz for PU, 2.5 to 4.5~kHz for CWRU), take the magnitude of
the analytic signal (the Hilbert envelope) to recover the modulation, decimate to
2~kHz, and normalize each window to zero mean and unit variance. For CWRU we
additionally normalize each raw window by its root-mean-square value beforehand,
which removes the overall energy level as a cue, since that energy varies with
load rather than with fault type. The CWRU healthy-baseline recordings are
sampled at 48~kHz whereas its fault recordings are 12~kHz~\citep{smith2015}; we
decimate the healthy recordings to 12~kHz before this pipeline so that all
classes pass through the same filters. The result is a 2000-sample envelope whose
spectrum shows peaks at the bearing fault frequencies.

\begin{figure}[t]
\centering
\resizebox{0.92\linewidth}{!}{%
\begin{tikzpicture}[
  >=Stealth, node distance=8mm, font=\small,
  s/.style={draw, rounded corners, align=center, text width=21mm, minimum height=12mm, inner sep=3pt, fill=blue!5},
  hz/.style={draw, rounded corners, align=center, text width=21mm, minimum height=11mm, inner sep=3pt, fill=gray!10},
  od/.style={draw, rounded corners, align=center, text width=21mm, minimum height=11mm, inner sep=3pt, fill=orange!14}
]
\node[s] (raw) {Raw window\\[2pt]\scriptsize 1\,s};
\node[s, right=of raw] (bp) {Band-pass\\[2pt]\scriptsize resonance band\\ 2--6 / 2.5--4.5\,kHz};
\node[s, right=of bp] (env) {Envelope\\[2pt]\scriptsize $\lvert x + i\,\mathcal{H}\{x\}\rvert$};
\node[s, right=of env] (dec) {Decimate to 2\,kHz\\[2pt]\scriptsize + normalize};
\node[hz, above right=4mm and 10mm of dec] (hzb) {Hz envelope\\[2pt]\scriptsize $L{=}2000$};
\node[od, below right=4mm and 10mm of dec] (odb) {Angle resample\\[2pt]\scriptsize 15\,rev $\times$ 128\\ $L{=}1920$};
\foreach \x/\y in {raw/bp, bp/env, env/dec} \draw[->] (\x) -- (\y);
\draw[->] (dec.east) -- (hzb.west);
\draw[->] (dec.east) -- (odb.west);
\end{tikzpicture}}

\vspace{4pt}
\includegraphics[width=0.94\linewidth]{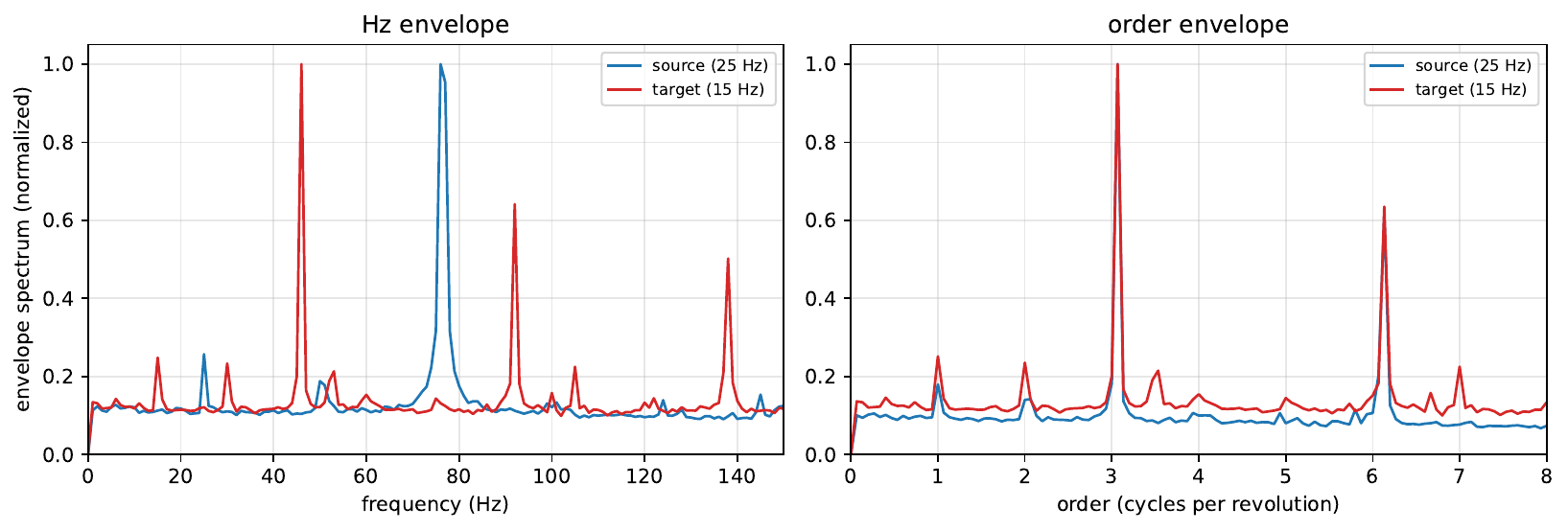}
\caption{Preprocessing. Top: each window is band-passed around the bearing
resonance, demodulated to its Hilbert envelope, decimated to 2\,kHz, and either
kept in the Hz domain or resampled against shaft angle to the order domain.
Bottom: envelope spectrum of the outer-race fault at the source and target
speeds. In the Hz domain the fault peak moves with speed; after order tracking it
falls at the same shaft order (near $3.05$ orders for the 6203).}
\label{fig:prep}
\end{figure}
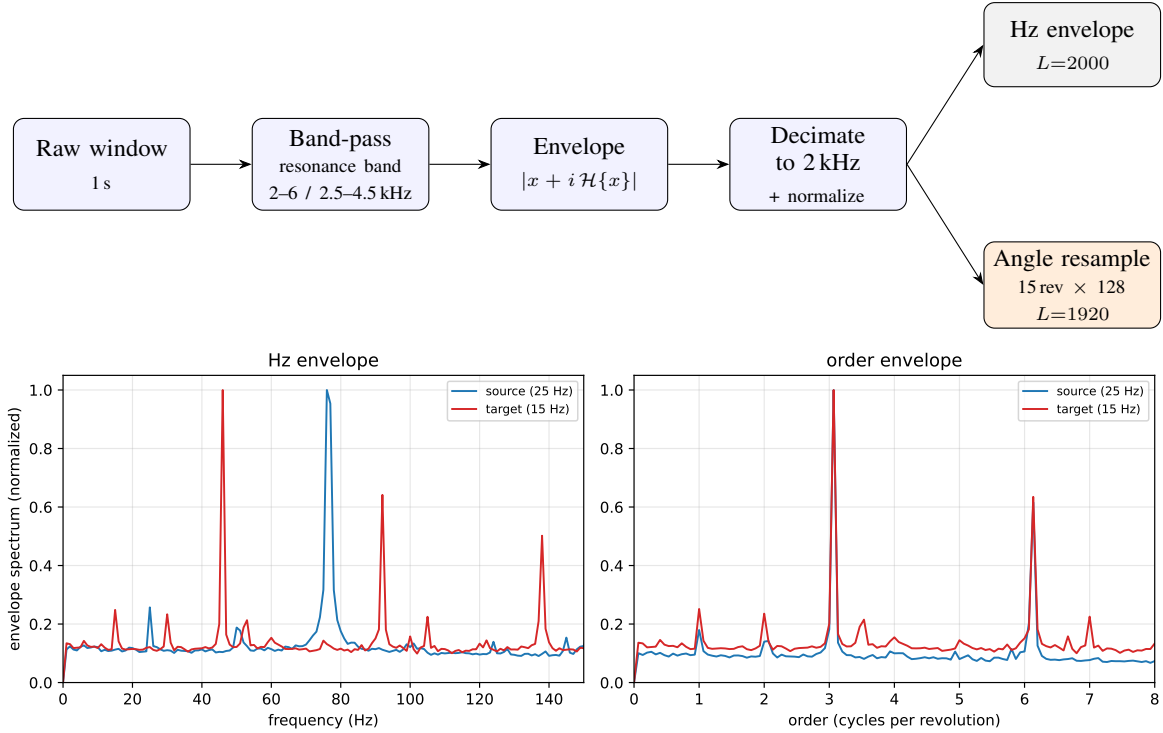

\paragraph{Order representation:}
The fault frequencies are fixed multiples of the shaft rate, so they move along
the frequency axis when the speed changes. To remove this movement we apply
computed order tracking. Because the shaft speed within each recording is
constant, we resample using the known nominal shaft rate of the condition rather
than an instantaneous tachometer signal: we place the envelope samples uniformly
in shaft angle rather than in time, so that a component at order $k$ completes $k$
cycles per shaft revolution regardless of speed. This assumes the target shaft
rate is known at test time, which holds for the fixed operating conditions
studied here. We fix each window to fifteen shaft revolutions sampled at 128
points per revolution, giving a 1920-sample signal in which a fault at a given
order falls at the same position regardless of speed. As a correctness check, the
outer-race peak lands near $3.05$ orders, the theoretical BPFO of the 6203, with
its first harmonic at $6.1$ (Figure~\ref{fig:prep}). Two caveats follow from the
fixed revolution count: the representations differ also in retained physical
duration (fifteen revolutions span $0.6$~s at $25$~Hz but $1.0$~s at $15$~Hz),
and at a fixed number of retained Fourier modes the two axes cover different
physical bandwidths (200 modes span $200$~Hz on the Hz axis but $333$~Hz at the
$25$~Hz shaft rate on the order axis). Section~\ref{sec:results} reports
bandwidth- and duration-matched controls that rule both differences out.

\paragraph{Held-out-unit protocol:}
We evaluate so that no physical bearing appears in both training and test. For
PU, each class is represented by three bearings, and we form three folds, each
holding out one bearing per class for testing and training on the remaining two.
CWRU provides a single recording per fault type and diameter, so we treat the
three fault diameters ($0.007$, $0.014$, $0.021$~inch) as the held-out units for
the fault classes and split the single healthy recording per load into three
contiguous segments; each fold then holds out one diameter and one healthy
segment. Because adjacent windows overlap by half, we drop the window that
straddles each segment boundary so that no test window shares samples with a
training window. Splitting a single recording remains a weaker guarantee than
holding out a distinct physical unit~\citep{vieira2025realistic}; PU, where every
class has three separate bearings, does not need this compromise. Two consequences follow and are visible in the results: the CWRU
folds test transfer to a fault \emph{geometry} unseen in training, and the
healthy class at the source condition has only five to six one-second training
windows. For validation and early stopping we reserve, on PU, recording
repetitions 15--17 of the training bearings (separate measurement runs, so no
window leakage), and on CWRU the last 20 percent of the windows of each training
recording in time order, stratified by class and with a one-window guard at the
boundary (recordings with fewer than eight windows contribute no validation
windows). Every non-oracle model is selected on this source-domain validation
split only; the oracle is selected on a target-domain validation split. No
labeled target window is used for model selection at any label fraction.

\paragraph{Domain adaptation setup:}
For a source condition $S$ and a target condition $T$, we train on the labeled
source data together with a fraction $f \in \{0, 0.03, 0.05, 0.10, 0.20, 0.30\}$
of labeled target data, and evaluate on the held-out target bearings. We report
two references: a source-only baseline (B1), trained on source labels alone,
which measures transfer without target information; and an oracle, trained on
target labels alone, which measures the supervised ceiling. On top of the
supervised loss we compare five choices of alignment on the unlabeled target
features: none, correlation alignment~\citep{sun2016coral}, domain-adversarial
training~\citep{ganin2016dann}, and two forms of maximum mean
discrepancy~\citep{gretton2012mmd}, one linear and one with a radial-basis kernel
on normalized features. Every configuration is run for three random seeds and
three folds; we report the mean and standard deviation over these nine runs and,
for the headline differences, paired tests over the nine matched runs
(Appendix~\ref{app:stats}). Test folds are class-balanced, so accuracy equals
balanced accuracy.

\paragraph{Architectures:}
We compare a Fourier Neural Operator and a convolutional network. The operator
lifts the input to sixteen channels and applies two spectral blocks, each
retaining the lowest 200 Fourier modes, followed by global average pooling and a
linear classifier. The convolutional network applies four convolutional blocks
with pooling, followed by global average pooling and a linear classifier. We
match the two at the feature layer, using a sixteen-dimensional feature vector in
both, so that the alignment losses, which act on this layer, operate on
representations of the same size. We do not match the total parameter count: the
operator uses about $207$k real-valued parameters (its spectral weights are
complex) and the convolutional network about $375$k, so the two are within a
factor of two.

\paragraph{Training:}
All models are trained with AdamW at a learning rate of $10^{-3}$ and weight
decay of $10^{-3}$, for up to 80 epochs with early stopping on the source-domain
validation split. We use a class-balanced cross-entropy loss with label
smoothing, mild amplitude jitter and additive noise for augmentation, and
gradient clipping. Hyperparameters and alignment-loss weights were chosen on the
operator and shared with the convolutional network. The operator reaches the
80-epoch budget in almost all runs, whereas the convolutional network typically
stops around epoch 50; both fit the source domain to a validation accuracy of
$1.00$ (Appendix~\ref{app:full}). Full settings are listed in
Appendix~\ref{app:hparams}. Code for the full pipeline, from raw recordings to
every table and figure, accompanies the submission.

\section{Experiments and Results}
\label{sec:results}
\begin{figure}[t]
\centering
\includegraphics[width=0.72\linewidth]{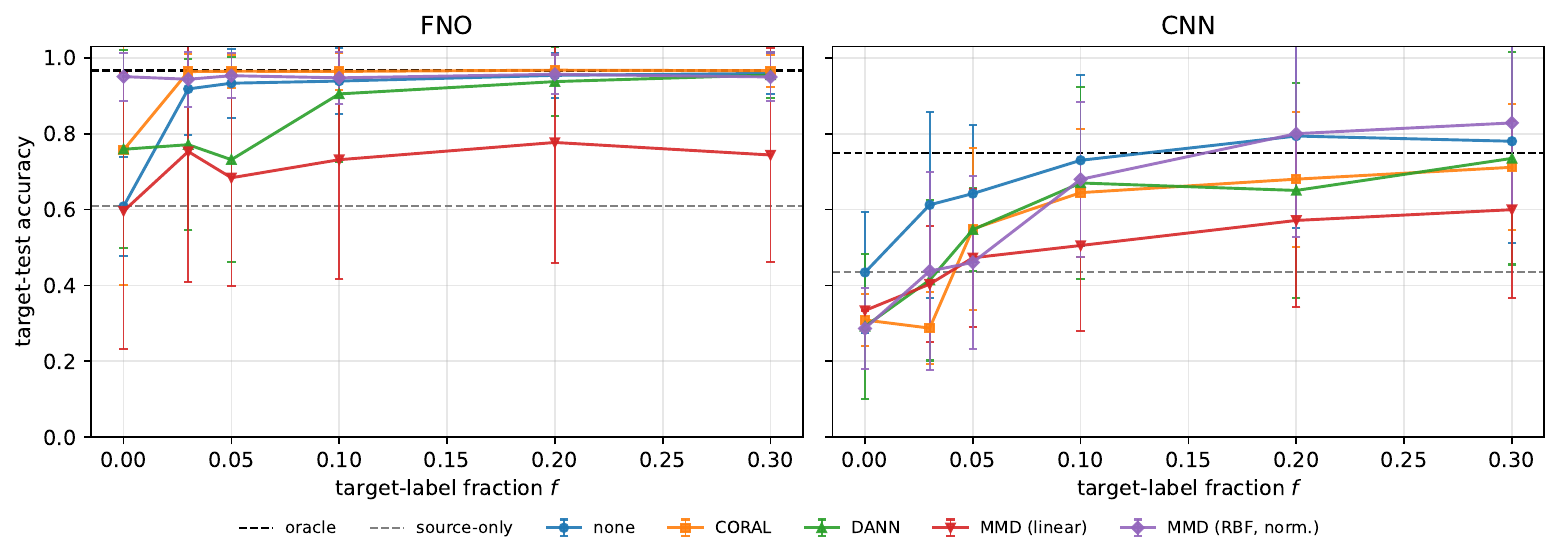}
\caption{Target accuracy against the target-label fraction $f$ on the PU speed
shift, in the order domain, with model selection on source validation only.
Dashed lines are the source-only baseline (gray) and the oracle (black). For the
operator, $f=0.03$ closes most of the gap and by $f\approx0.2$ the stable
alignment methods are indistinguishable. Error bars: one standard deviation over
the nine runs.}
\label{fig:accf}
\end{figure}

Table~\ref{tab:main} reports source-only transfer (B1) and the target-supervised
ceiling (oracle) on PU under both representations, Table~\ref{tab:aligners}
reports the five alignment choices without target labels, and
Figure~\ref{fig:accf} the label-fraction curves.

\paragraph{Order tracking recovers source-only transfer for the neural operator:}
For the Fourier Neural Operator, moving from the frequency-domain envelope to the
order-domain envelope raises source-only accuracy from $0.36$ to $0.61$ on the PU
speed shift ($+0.25$, eight of nine matched runs, Wilcoxon $p=0.012$). The
per-class breakdown shows the mechanism directly. In the frequency domain the
source-only operator labels almost every target window healthy (per-class recall
$1.00 / 0.03 / 0.04$ for healthy / outer / inner): at 900~rpm the fault peaks
fall at frequency bins the source model learned to ignore. In the order domain
the fault classes are recovered ($0.74$ outer, $0.84$ inner) but healthy drops to
$0.24$: the peaks now sit where the classifier looks, yet the overall
peak-versus-no-peak decision still shifts with speed. Unsupervised alignment
removes exactly this remaining shift, reaching $1.00 / 0.97 / 0.88$ (below). The
convolutional network changes little between the two representations on this
shift ($+0.06$, four of nine runs, $p=0.65$) and stays near chance in both.

\paragraph{The order axis is harmless where the shift is not spectral:}
Fault frequencies move only when the operating change alters shaft speed. The PU
torque and radial-force shifts vary the load at a fixed speed, so the peaks do
not move; there the frequency-domain baseline is already near ceiling ($0.95$ and
$0.93$), and order tracking neither helps nor hurts ($0.94$ and $0.93$): the
resampling is benign when there is nothing spectral to correct.

\paragraph{Unsupervised alignment works only in the right representation:}
Table~\ref{tab:aligners} gives every alignment loss at $f=0$. For the operator,
all four stable losses transfer better in the order domain than in the frequency
domain, and the best of them, normalized RBF-MMD, reaches $0.95 \pm 0.06$ on the
PU speed shift in the order domain against $0.71 \pm 0.33$ in the frequency
domain; the order-domain value is within $0.02$ of the oracle and at least $0.86$
on every one of the nine held-out-bearing runs (per fold $1.00 / 0.87 / 0.98$).
The improvement over the source-only order model is uniform ($+0.34$, nine of nine runs, $p=0.004$),
whereas the improvement over the frequency-domain aligned model is concentrated
in the fold where the frequency-domain model fails (bootstrap 95 percent
interval $[+0.05, +0.42]$, four of nine runs, $p=0.16$): in two folds the
frequency-domain model already reaches $1.00$ for some seeds, so the order axis
there buys reliability rather than mean accuracy, reducing the standard deviation
from $0.33$ to $0.06$. The same losses applied to the convolutional network help
in neither domain, and in the order domain they degrade it ($0.43$ source-only
to $0.29$--$0.33$); the loss weights were selected on the operator and shared
(Section~\ref{sec:discussion}). This is the sense in which the representation, not the
alignment method, decides whether adaptation helps: the alignment objective is
identical across the two columns of Table~\ref{tab:aligners}, and only the input
axis differs.

\paragraph{A small amount of labeled target data closes the remaining gap:}
Figure~\ref{fig:accf} shows target accuracy against the target-label fraction $f$
on the PU speed shift in the order domain, with every run selected on source
validation only. For the operator, $f=0.03$ already brings the no-alignment model
from $0.61$ to $0.92$ and correlation alignment to $0.96$; by $f=0.10$ three of
the four stable alignment choices lie within $0.03$ of the oracle, and by
$f=0.20$ all four do.
Linear MMD remains unstable at every fraction and is shown for completeness.
The value of unsupervised alignment is thus concentrated at
$f=0$; once a few target labels are available, the choice of alignment method has
little effect. The convolutional network improves more slowly, and even at $f=0.30$ its
best variant reaches only $0.83$.

\paragraph{The neural operator is the stronger learner for this signal:}
In the order domain the operator exceeds the convolutional network on the PU
speed shift both without target labels ($0.61$ against $0.43$; $+0.17$, eight of
nine runs, $p=0.008$) and with them (oracle $0.97$ against $0.75$). We do not
claim this follows from the convolutional network responding to local time
patterns, since we did not isolate that mechanism; both models fit the source
domain to a validation accuracy of $1.00$, so the gap is a transfer gap, and the
alignment weights were selected on the operator and shared rather than retuned.

\paragraph{Controls:}
Five controls verify that the order advantage is what it appears to be; full
versions are in Appendices~\ref{app:cnnsearch} and~\ref{app:controls}. (i)~A
27-configuration search over the convolutional network's width, depth, and
learning rate leaves its oracle at $0.64$--$0.80$ and its source-only transfer
at $0.26$--$0.44$, never approaching the operator's $0.97$ and $0.61$: the
architecture gap is not an artifact of shared hyperparameters. (ii)~A two-layer
MLP on the log-magnitude spectrum reproduces the direction of the
representation effect (source-only $0.54$ order against $0.42$ Hz) but not its
size or reliability ($0.85\pm0.16$ under RBF-MMD at $f{=}0$ against the
operator's $0.95\pm0.06$; oracle $0.88$ against $0.97$), so a global spectral
view accounts for only part of the gain. (iii)~The advantage survives bandwidth
matching ($0.61$ against $0.43$ at matched $333$~Hz bands) and holds against
the full Hz spectrum ($0.44$). (iv)~It is not a duration effect: a fixed
$0.6$~s crop of the Hz window leaves source-only accuracy at $0.37$, whereas a
fifteen-revolution crop, which order-aligns the DFT without any
resampling, recovers $0.63$, matching the order value; the operator needs only
a whole number of shaft revolutions in the window. (v)~Zeroing the input modes
at the fault orders confirms the operator reads them: the BPFO band alone
carries the outer-race class for the source-only model (recall $0.74$ to
$0.33$; to $0.00$ in two of three folds), the effects are class-specific, and
control bands are inert (Table~\ref{tab:ablation}).

\paragraph{Secondary dataset, unseen fault geometry defeats even the oracle:}
Table~\ref{tab:cwru} reports CWRU per fold, because the folds differ in kind. In
fold~1 (train on $0.014$ and $0.021$~inch faults, test on $0.007$~inch) both
models transfer across the $3.7$ percent load-speed shift: the convolutional
network reaches $1.00$ in the frequency domain and $0.99$ in the order domain,
and the operator reaches $0.91$ in the order domain and $0.52$ in the frequency
domain, where its failure is confined to the inner-race class (recall $0.00$). This deficit is not a
bandwidth effect: doubling the frequency-domain band to 400 retained modes
leaves the fold-1 inner-race recall at $0.00$ for the source-only model, while
the oracle at the same band reaches $0.82$, so the information is present in the
band and the failure is one of transfer. Nor is it a duration effect: a $0.5$~s
crop of the frequency-domain window leaves fold-1 source-only accuracy at $0.52$
(Appendix~\ref{app:controls}). The angle resampling
evidently removes enough of the small load-induced variation for the inner-race
signature to carry across the unseen diameter, where the raw frequency axis does
not. Folds~2 and~3 hold out the $0.014$ and $0.021$~inch
faults and are a different test: in fold~2 the held-out outer-race class has
recall $0.09$ for the source model and $0.00$ for the oracle in both
representations, and in fold~3 the held-out inner-race class has recall $0.01$
for the source model and $0.44$ for the order-domain oracle. A model trained on
two fault diameters, with or without target labels, does not recognize the third.
This is a failure of generalization to unseen fault geometry, not of adaptation
across load, and a window-level split would have hidden it entirely.

\begin{table}[t]
\centering
\caption{PU: source-only transfer (B1) and target-supervised ceiling (oracle)
under the frequency (Hz) and order representations. Order tracking lifts the
neural operator on the shift that moves the fault frequencies and is benign
elsewhere; the convolutional network is largely unchanged. Means over nine runs
(three seeds $\times$ three held-out-bearing folds); standard deviations and
source-validation accuracy in Appendix~\ref{app:full}. $\Delta$ is order minus
Hz.}
\label{tab:main}
\small
\begin{tabular}{llccccc}
\toprule
Model & Shift & B1 (Hz) & B1 (order) & $\Delta$ & Oracle (Hz) & Oracle (order) \\
\midrule
FNO & speed   & 0.36 & 0.61 & $+0.25$ & 0.97 & 0.97 \\
FNO & torque  & 0.95 & 0.94 & $-0.01$ & 0.95 & 0.95 \\
FNO & load    & 0.93 & 0.93 & $\phantom{+}0.00$ & 0.97 & 1.00 \\
\midrule
CNN & speed   & 0.38 & 0.43 & $+0.06$ & 0.65 & 0.75 \\
CNN & torque  & 0.81 & 0.91 & $+0.10$ & 0.81 & 0.83 \\
CNN & load    & 0.78 & 0.78 & $\phantom{+}0.00$ & 0.78 & 0.69 \\
\bottomrule
\end{tabular}
\end{table}

\begin{table}[t]
\centering
\caption{Unsupervised alignment ($f=0$) on the PU speed shift, mean $\pm$
standard deviation over nine runs. For the operator, every stable loss transfers
better in the order domain, and normalized RBF-MMD reaches the oracle only
there. For the convolutional network no loss helps in either domain. Source-only
(no alignment) and oracle rows are references.}
\label{tab:aligners}
\small
\begin{tabular}{llcc}
\toprule
Model & Alignment ($f{=}0$) & Hz & Order \\
\midrule
\multirow{6}{*}{FNO}
 & none (source-only)   & $0.36\pm0.04$ & $0.61\pm0.13$ \\
 & CORAL                & $0.63\pm0.27$ & $0.76\pm0.36$ \\
 & DANN                 & $0.55\pm0.16$ & $0.76\pm0.26$ \\
 & MMD (linear)         & $0.44\pm0.14$ & $0.60\pm0.36$ \\
 & MMD (RBF, normalized)& $0.71\pm0.33$ & $\mathbf{0.95\pm0.06}$ \\
 & oracle               & $0.97\pm0.03$ & $0.97\pm0.04$ \\
\midrule
\multirow{6}{*}{CNN}
 & none (source-only)   & $0.38\pm0.16$ & $0.43\pm0.16$ \\
 & CORAL                & $0.41\pm0.22$ & $0.31\pm0.07$ \\
 & DANN                 & $0.39\pm0.24$ & $0.29\pm0.19$ \\
 & MMD (linear)         & $0.47\pm0.29$ & $0.33\pm0.00$ \\
 & MMD (RBF, normalized)& $0.34\pm0.20$ & $0.29\pm0.11$ \\
 & oracle               & $0.65\pm0.16$ & $0.75\pm0.13$ \\
\bottomrule
\end{tabular}
\end{table}

\begin{table}[t]
\centering
\caption{CWRU per fold (neural operator; mean $\pm$ std over three loads
$\times$ three seeds). Each fold holds out one fault diameter for testing.
Fold~1 is a transfer test across the load shift: the operator's
frequency-domain deficit there is confined to the inner-race class and persists
at doubled bandwidth (see text). Folds~2 and~3 are unseen-geometry tests that the
target-supervised oracle also fails on the held-out class. Zero standard
deviations are real: on these small test folds every run makes the same
predictions (Section~\ref{sec:discussion}).}
\label{tab:cwru}
\footnotesize
\setlength{\tabcolsep}{3pt}
\begin{tabular}{lcccccc}
\toprule
 & \multicolumn{2}{c}{B1} & \multicolumn{2}{c}{Oracle} & \multicolumn{2}{c}{Held-out class recall (outer / inner)} \\
\cmidrule(lr){2-3}\cmidrule(lr){4-5}\cmidrule(lr){6-7}
Fold (held-out diameter) & Hz & Order & Hz & Order & B1 (order) & Oracle (order) \\
\midrule
1 ($0.007$ in.) & $0.52\pm0.06$ & $0.91\pm0.10$ & $0.87\pm0.20$ & $1.00\pm0.00$ & $1.00$ / $0.95$ & $1.00$ / $1.00$ \\
2 ($0.014$ in.) & $0.37\pm0.20$ & $0.61\pm0.06$ & $0.47\pm0.19$ & $0.57\pm0.00$ & $0.09$ / $1.00$ & $0.00$ / $1.00$ \\
3 ($0.021$ in.) & $0.55\pm0.08$ & $0.50\pm0.08$ & $0.86\pm0.22$ & $0.75\pm0.24$ & $0.99$ / $0.01$ & $0.98$ / $0.44$ \\
\bottomrule
\end{tabular}
\end{table}

\section{Discussion and Limitations}
\label{sec:discussion}
\paragraph{The protocol exposes failures that random splits hide:}
Held-out units reveal per-unit failures that pooled accuracy conceals. On PU, the
operator's source-only accuracy in the order domain is $0.68\pm0.03$,
$0.49\pm0.18$, and $0.65\pm0.00$ by fold; in the second fold the held-out
outer-race bearing KA03 is recognized only $0.25$ of the time and the inner-race
bearing KI03 only $0.53$. On CWRU the same effect is stronger and independent of labels:
the $0.014$~inch outer-race fault is not recognized by any model trained on the
other two diameters, oracle included (Table~\ref{tab:cwru}). A split that let windows from these
units into training would hide both failures and overstate generalization to
new hardware. Smith and Randall~\citep{smith2015} note that several CWRU $0.014$~inch
records are difficult or impossible to diagnose with conventional envelope
analysis; our result is the learned-model counterpart.

\paragraph{Order tracking is a tool, not a contribution:}
Removing speed dependence through order tracking is standard in
diagnostics~\citep{randall2011}, and has been combined with deep networks and with
domain adaptation before~\citep{ji2021order, xu2023tco}. We use it here as a
controlled way to change the representation while holding everything else fixed,
which is what lets us attribute the operator's gains to the representation rather
than to the alignment loss. The matched-band and matched-duration controls make
this attribution safe: either difference could by itself have produced the
gains, and Appendix~\ref{app:controls} rules both out on both datasets.

\paragraph{Anomalies and scope:}
On CWRU folds~2 and~3 and on the PU load shift for the convolutional network,
the oracle sits at or below the source-only baseline. On CWRU this is expected: an oracle trained on two fault diameters need
not extrapolate to the third any better than a source model does, so the ceiling
itself is not tight on those folds. Linear maximum mean discrepancy at weight $100$ is unstable and collapses the
operator to a single class on some runs; we report it but do not recommend it. The
alignment weights and optimizer settings were chosen on the operator and shared
with the convolutional network, so the architecture comparison is an observation
rather than a tuned benchmark. Our order tracking assumes the target shaft rate is
known and constant within a recording, which holds here but would need a
tachometer or tacholess estimate under run-up or coast-down; estimating the speed from
the vibration signal itself is well studied~\citep{lu2019tacholess}, and a
measured sensitivity analysis shows the representation needs the assumed rate to
about one percent, within tacholess accuracy (Appendix~\ref{app:controls},
Table~\ref{tab:spderr}). On CWRU the single healthy recording per load leaves
five to six healthy training windows at the source
condition once segments are held out, and each fold is tested on one recording
per fault class, so its accuracies are coarsely quantized and its error bars wide.
Finally, we study a one-dimensional envelope representation; whether the
representation-over-alignment conclusion holds for two-dimensional time-frequency
inputs is left to future work.

\section{Conclusion}
\label{sec:conclusion}
We revisited domain adaptation for bearing fault diagnosis under a
held-out-unit protocol, where source-only transfer proves far weaker than the
near-perfect numbers usually reported. Using order tracking as a controlled
change of representation, we found that a Fourier Neural Operator's transfer
improves on the shift that moves the fault frequencies, and that the same
representation is what lets unsupervised alignment close the gap at all, while
a convolutional network of the same feature size is almost unaffected: the
input representation, not the alignment method, decides whether adaptation
helps. A small fraction of target labels closes most of the remaining gap, and
the same protocol reveals that no model, target-supervised included, recognizes
a fault geometry it has not seen. Practitioners should ask whether a shift
moves the fault frequencies, choose the representation accordingly, control
what the change also changes, and evaluate on unit-disjoint splits.


\small
\bibliographystyle{plainnat}

\section{Appendix}
\subsection{Full PU results with standard deviations}
\label{app:full}
Table~\ref{tab:full} repeats Table~\ref{tab:main} with standard deviations over
the nine runs (three seeds, three folds) and the source-validation accuracy of
the selected checkpoint, which is $1.00$ for every source-only model, so all
transfer gaps are gaps in transfer, not in fitting.

\begin{table}[h]
\centering
\caption{PU source-only (B1) and oracle accuracy, mean $\pm$ standard deviation
over nine runs, under both representations, with the source-validation accuracy
of the source-only model.}
\label{tab:full}
\small
\begin{tabular}{llccccc}
\toprule
Model & Shift & B1 (Hz) & B1 (order) & Oracle (Hz) & Oracle (order) & Src.\ val.\ (B1) \\
\midrule
FNO & speed   & $0.36\pm0.04$ & $0.61\pm0.13$ & $0.97\pm0.03$ & $0.97\pm0.04$ & $1.00$ \\
FNO & torque  & $0.95\pm0.11$ & $0.94\pm0.10$ & $0.95\pm0.11$ & $0.95\pm0.10$ & $1.00$ \\
FNO & load    & $0.93\pm0.13$ & $0.93\pm0.13$ & $0.97\pm0.06$ & $1.00\pm0.01$ & $1.00$ \\
\midrule
CNN & speed   & $0.38\pm0.16$ & $0.43\pm0.16$ & $0.65\pm0.16$ & $0.75\pm0.13$ & $1.00$ \\
CNN & torque  & $0.81\pm0.17$ & $0.91\pm0.12$ & $0.81\pm0.16$ & $0.83\pm0.13$ & $1.00$ \\
CNN & load    & $0.78\pm0.16$ & $0.78\pm0.14$ & $0.78\pm0.13$ & $0.69\pm0.08$ & $1.00$ \\
\bottomrule
\end{tabular}
\end{table}

\subsection{Paired statistics for the headline differences}
\label{app:stats}
Each difference is computed on the nine matched (fold, seed) runs of the PU speed
shift. We report the mean difference, a paired-bootstrap 95 percent interval
(10{,}000 resamples), the number of runs in which the first configuration wins,
and the two-sided Wilcoxon signed-rank $p$-value.

\begin{table}[h]
\centering
\small
\begin{tabular}{lcccc}
\toprule
Comparison & Mean $\Delta$ & 95\% interval & Wins & $p$ \\
\midrule
FNO source-only: order $-$ Hz                    & $+0.25$ & $[+0.15, +0.33]$ & 8/9 & $0.012$ \\
CNN source-only: order $-$ Hz                    & $+0.06$ & $[-0.07, +0.18]$ & 4/9 & $0.65$ \\
FNO order: RBF-MMD ($f{=}0$) $-$ source-only      & $+0.34$ & $[+0.28, +0.40]$ & 9/9 & $0.004$ \\
FNO RBF-MMD ($f{=}0$): order $-$ Hz               & $+0.24$ & $[+0.05, +0.42]$ & 4/9 & $0.16$ \\
FNO DANN ($f{=}0$): order $-$ Hz                  & $+0.21$ & $[+0.03, +0.40]$ & 5/9 & $0.11$ \\
FNO CORAL ($f{=}0$): order $-$ Hz                 & $+0.13$ & $[-0.06, +0.33]$ & 4/9 & $0.44$ \\
Order source-only: FNO $-$ CNN                   & $+0.17$ & $[+0.08, +0.27]$ & 8/9 & $0.008$ \\
\midrule
FNO B1: Hz $0.6$\,s $-$ Hz $1.0$\,s              & $+0.01$ & $[-0.02, +0.06]$ & 3/9 & $0.74$ \\
FNO B1: Hz $15$\,rev $-$ Hz $1.0$\,s             & $+0.28$ & $[+0.17, +0.38]$ & 8/9 & $0.008$ \\
FNO B1: Hz $15$\,rev $-$ order                   & $+0.03$ & $[-0.01, +0.08]$ & 3/9 & $0.38$ \\
\bottomrule
\end{tabular}
\caption{Paired differences over the nine matched (fold, seed) runs of the PU speed shift.
The last three rows are the duration control (Appendix~\ref{app:controls}): matching physical
duration alone does not change transfer, while matching shaft revolutions recovers the
order-domain result.}
\end{table}

The order-versus-Hz comparisons under alignment show few ``wins'' because in
folds~1 and~3 the frequency-domain aligned model already reaches $1.00$ for some
seeds; the order axis there removes the failing seeds and the failing fold rather
than raising an already-perfect mean, which is why its effect appears in the
standard deviation ($0.33$ to $0.06$) and the run minimum ($0.33$ to $0.86$) more
than in the paired count.

\subsection{Convolutional-network hyperparameter search}
\label{app:cnnsearch}
We swept the convolutional network's base width, depth, and learning rate on the
PU speed shift in the order domain ($27$ configurations, three seeds $\times$
three held-out-bearing folds each, B1 and oracle only). Because every
configuration reaches a source-validation accuracy of $0.99$--$1.00$ (never
below $0.97$ in any run; Table~\ref{tab:cnnsearch}), source validation, the
only label-free selection signal available, cannot distinguish them, which is
itself the point: no width/depth/learning-rate choice converts the convolutional
network into one that transfers. The best oracle over the whole grid ($0.80$,
from the smallest-width deepest network) still trails the operator's $0.97$ by
$0.17$, with a standard deviation of $0.20$, and no configuration lifts
source-only transfer beyond $0.44$, far below the operator's $0.61$.

\begin{table}[h]
\centering
\caption{Convolutional-network grid search, PU speed shift, order domain. Oracle
and source-only (B1) accuracy, mean $\pm$ std over nine runs, with the mean
source-validation accuracy of each configuration. The operator is shown for
reference.}
\label{tab:cnnsearch}
\small
\begin{tabular}{lccc}
\toprule
Configuration & Src.\ val. & Oracle & Source-only (B1) \\
\midrule
CNN, best oracle (width 32, depth 5) & $0.99$ & $0.80\pm0.20$ & $0.40\pm0.22$ \\
CNN, default (width 48, depth 4) & $1.00$ & $0.75\pm0.13$ & $0.43\pm0.16$ \\
CNN, worst oracle (width 64, depth 4, lr $3{\times}10^{-3}$) & $0.99$ & $0.64\pm0.19$ & $0.39\pm0.16$ \\
\midrule
FNO (reference) & $1.00$ & $\mathbf{0.97\pm0.04}$ & $\mathbf{0.61\pm0.13}$ \\
\bottomrule
\end{tabular}
\end{table}

\subsection{Hyper-parameters and configuration}
\label{app:hparams}
The operator uses a lifting width of 16, two spectral blocks with hidden width 16
and 200 retained Fourier modes, group normalization, and dropout 0.3 ($206{,}807$
real-valued parameters; its spectral weights are complex, so a count that treats
each complex number as one parameter reports $103{,}189$). The convolutional
network uses four blocks with base width 48, a sixteen-dimensional feature layer,
and dropout 0.3 ($373{,}939$ parameters). Both are trained with AdamW (learning
rate $10^{-3}$, weight decay $10^{-3}$) for up to 80 epochs, batch size 64, with
early stopping (patience 20) on a source-domain validation split. We use
class-balanced cross-entropy with label smoothing $0.05$, and augment with
amplitude jitter of $\pm10$ percent and additive Gaussian noise of standard
deviation $0.05$, with gradient clipping at norm $1.0$. Alignment-loss weights are
50 for correlation alignment, 1 for domain-adversarial training with a ramp on the
reversal strength, 100 for linear maximum mean discrepancy, and 5 for the
radial-basis variant; the radial-basis kernel uses bandwidths $\gamma \in
\{0.5, 1, 2, 4\}$ on $\ell_2$-normalized features. All experiments run on a single NVIDIA RTX 4090; a full 864-run sweep takes about $4.3$ GPU-hours on PU and $2.3$ on CWRU,
with $3$--$12$~s per training run.

\subsection{Additional controls: spectrum input, bandwidth, duration, mode
ablation, and speed error}
\label{app:controls}
This appendix gives the full versions of the controls summarized in
Section~\ref{sec:results}.

\paragraph{Spectrum-input control:}
To test whether the order-domain gains
require the neural operator or only \emph{a} global spectral view, we repeat
the $f{=}0$ comparison with a two-layer MLP that reads the log-magnitude
spectrum of its input and shares the same sixteen-dimensional feature layer
and alignment losses. The spectral MLP reproduces the direction of the
representation effect (source-only $0.54$ in the order domain against $0.42$
in Hz) but not its size or reliability: under normalized RBF-MMD at $f{=}0$
it reaches $0.85\pm0.16$ against the operator's $0.95\pm0.06$, and its
target-supervised ceiling is $0.88$ against the operator's $0.97$ (paired
over the nine matched runs, $+0.09$, eight of nine, $p=0.008$). A global
spectral view thus accounts for part of the gain; the operator's spectral
parametrization contributes beyond it.

\paragraph{Bandwidth control:}
Because 200 retained modes cover $200$~Hz on the Hz axis but $333$~Hz on the
order axis at the source speed, we repeat the comparison with matched bands and
with the full spectrum retained on both axes. The order advantage survives
bandwidth matching: with both axes restricted to $333$~Hz, source-only accuracy
is $0.43$ in the frequency domain (333 modes) against $0.61$ in the order domain
(200 modes), and with both restricted to $200$~Hz it is $0.36$ against $0.63$
(120 modes). Even with the full spectrum retained, the frequency-domain model
reaches only $0.44$, still well below the order model at any band. Under
alignment the pattern is the same: normalized RBF-MMD at $f{=}0$ reaches
$0.90\pm0.20$ in the frequency domain at 333 modes against $0.95\pm0.06$ in the
order domain, the difference again lying in reliability across folds. Retaining
more modes does help \emph{within} the order domain (source-only $0.80$ at the
full 961 modes), so bandwidth has a real secondary effect, but it acts on top of
the coordinate change rather than explaining it.

\paragraph{Duration control:}
Fixing the window at 15 revolutions makes the order representation span
$0.6$~s of a $25$~Hz source window but the full $1.0$~s of a $15$~Hz target
window, so the two representations differ in retained duration as well as in
coordinate. Two crops of the frequency-domain window separate these. Cropping
every window to a fixed $0.6$~s at both speeds---the physical span of the source
order window---leaves source-only accuracy at $0.37\pm0.05$, indistinguishable
from the $1.0$~s value ($0.36$; paired $\Delta=+0.01$, $p=0.74$): the duration
difference alone does not move transfer. Cropping instead to exactly fifteen
revolutions, which places a component of order $k$ at DFT bin $k/15$ at both
speeds and so order-aligns the spectrum without any angle resampling, raises the
operator to $0.63\pm0.15$ ($+0.28$ over the $1.0$~s frequency baseline, eight of
nine runs, $p=0.008$), matching the order-domain value ($0.61$; $\Delta=+0.03$,
$p=0.38$), while the convolutional network is unchanged at $0.43$. The operator
therefore needs only that the window contain a whole number of shaft revolutions
for its transform to see the fault orders at fixed positions; the resampling adds
nothing beyond this, and the order advantage is a coordinate effect, not a
duration one. On CWRU the same crop rules out a duration explanation there too:
on the load-transfer fold, where order tracking lifts source-only accuracy from
$0.52$ to $0.91$, cropping the frequency-domain window to the $0.5$~s order span
leaves it at $0.52$.

\paragraph{Mode ablation at the fault orders:}
To test whether the operator reads the fault orders the representation fixes, we
zero at inference the input Fourier modes in a $\pm 0.15$-order band around each
fault order (BPFO $3.05$, its harmonic $6.10$, BPFI $4.95$, $2{\times}$BPFI) and
in two control bands where no fault energy sits ($1.5$--$2.0$ and $7.0$--$7.5$
orders). The effects are class-specific and land exactly where the kinematics
predict: removing the BPFO band costs the source-only model $0.13$ in accuracy,
all of it in the outer-race class (recall $0.74$ to $0.33$; in two of the three
folds the held-out outer bearing falls to $0.00$), while for the oracle,
removing the BPFO band reduces only outer-race recall ($0.98$ to $0.89$) and
removing the BPFI band only inner-race recall ($1.00$ to $0.84$). Every
control-band ablation changes accuracy by less than $0.01$. That the inner class
survives removal of its fundamental is itself the expected physics: an
outer-race fault is fixed relative to the load zone and its envelope energy
concentrates at the BPFO harmonics, whereas an inner-race fault is modulated at
the shaft rate and its energy spreads into sidebands around
BPFI~\citep{randall2011}, so no single narrow band is critical
(Table~\ref{tab:ablation}).

\paragraph{Speed-error sensitivity:}
We measured how much the known-rate assumption matters: a resampling error that
displaces every order by one percent reduces source-only order-domain accuracy
from $0.61$ to $0.51$, and a two percent displacement removes the order
advantage entirely ($0.35$, the frequency-domain level), whereas a perturbation
that changes only the sampling density and leaves the orders in place has no
effect ($0.62$ at up to five percent). The representation thus requires the
shaft rate to roughly one percent, well within the reported accuracy of
tacholess estimation~\citep{lu2019tacholess} (Table~\ref{tab:spderr}).

\begin{table}[H]
\centering\small
\begin{tabular}{lcccc}
\toprule
 & \multicolumn{2}{c}{Source-only (B1)} & \multicolumn{2}{c}{Oracle} \\
Ablated band (orders) & acc & outer / inner & acc & outer / inner \\
\midrule
none                        & $0.61$ & $0.74$ / $0.84$ & $0.97$ & $0.98$ / $0.92$ \\
BPFO $2.90$--$3.20$         & $0.48$ & $0.33$ / $0.87$ & $0.96$ & $0.89$ / $1.00$ \\
$2{\times}$BPFO $5.95$--$6.25$ & $0.60$ & $0.72$ / $0.84$ & $0.97$ & $0.98$ / $0.92$ \\
BPFI $4.80$--$5.10$         & $0.60$ & $0.74$ / $0.83$ & $0.94$ & $0.98$ / $0.84$ \\
$2{\times}$BPFI $9.75$--$10.05$ & $0.61$ & $0.74$ / $0.84$ & $0.97$ & $0.98$ / $0.91$ \\
control $1.50$--$2.00$      & $0.61$ & $0.74$ / $0.84$ & $0.96$ & $0.98$ / $0.89$ \\
control $7.00$--$7.50$      & $0.61$ & $0.74$ / $0.85$ & $0.97$ & $0.98$ / $0.92$ \\
\bottomrule
\end{tabular}
\caption{Mode ablation on the PU speed shift (order domain, neural operator; mean over
nine runs). Zeroing a fault-order band affects only its own class; control bands are
inert. The outer class depends on the BPFO fundamental while the inner class survives
removal of its fundamental, matching the concentrated-versus-sideband structure of the
two fault types.}
\label{tab:ablation}
\end{table}
 
\begin{table}[H]
\centering\footnotesize
\setlength{\tabcolsep}{2.5pt}
\begin{tabular}{lccccccc}
\toprule
Assumed-rate error & $-5\%$ & $-2\%$ & $-1\%$ & $0$ & $+1\%$ & $+2\%$ & $+5\%$ \\
\midrule
B1 accuracy & $0.34\pm0.10$ & $0.35\pm0.04$ & $0.51\pm0.14$ & $0.61\pm0.12$ & $0.62\pm0.11$ & $0.62\pm0.10$ & $0.62\pm0.11$ \\
\bottomrule
\end{tabular}
\caption{Sensitivity of source-only order-domain accuracy to an erroneous assumed shaft
rate (PU speed shift, neural operator, mean $\pm$ std over nine runs). Negative
perturbations displace every order by the stated fraction and are the faithful model of
a real rate error of either sign; positive perturbations change only the sampling
density and leave the orders on their bins, serving as a control.}
\label{tab:spderr}
\end{table}

\end{document}